\documentclass{esannV2}
\usepackage[dvips]{graphicx}
\usepackage[latin1]{inputenc}
\usepackage{amssymb,amsmath,array}
\usepackage{paralist}

\usepackage{subfigure} 

\usepackage{algorithmic}
\usepackage{algorithm}
\usepackage{marginnote}
\usepackage{latexsym}
\usepackage{graphicx}
\usepackage{multicol}
\usepackage{xcolor}
\usepackage{url}
\usepackage{booktabs}
\usepackage{hyperref}
\usepackage{dcolumn}

\newcommand{\sign}{\mathop{\mathrm{sign}}}

\newcommand{\R}{\mathbb{R}}

\begin{document}
	\title{Fast model selection\\ by limiting SVM training times}

\author{Ayd{\i}n Demircio\u{g}lu$^1$, Daniel Horn$^2$, Tobias Glasmachers$^1$, Bernd Bischl$^3$, Claus Weihs$^2$ 
%
\thanks{We acknowledge support by the Mercator Research Center Ruhr,
	under grant Pr-2013-0015 \textit{Support-Vektor-Maschinen f{\"u}r extrem gro{\ss}e Datenmengen}}
%
\vspace{.3cm}\\
%
1- Ruhr-Universit\"at Bochum - Institut f\"ur Neuroinformatik\\
44780 Bochum, Germany 
%
\vspace{.1cm}\\
2- Technische Universit\"at Dortmund - Fakult\"at Statistik\\
44221 Dortmund, Germany 
%
\vspace{.1cm}\\
3- Ludwig-Maximilians-Universit\"at M\"unchen - Institut f\"ur Statistik\\
80539 M\"unchen, Germany}

\maketitle

\begin{abstract}
Kernelized Support Vector Machines (SVMs) are among
the best performing supervised learning methods.
But for optimal predictive performance, 
time-consuming parameter tuning is crucial,
which impedes application.
To tackle this problem, the classic 
model selection procedure based on grid-search and cross-validation 
was refined, e.g. by data subsampling and
direct search heuristics.
Here we focus on a different aspect, the stopping
criterion for SVM training. 
We show that by limiting the training time
given to the SVM solver during parameter tuning 
 we can reduce model selection times by an order of magnitude.
\end{abstract}

\section{Introduction}

One of the standard classifiers for solving 
machine learning problems are kernelized Support Vector Machines (SVMs).
While they yield excellent performance, they suffer
from two problems: On the one hand, solving the
underlying optimization problem to a fixed accuracy has at least
quadratic complexity in the number of training points.
On the other hand, both, the regularization parameter
as well as the kernel parameters need heavy, problem-specific tuning.

The first problem is addressed by a multitude of approximate solvers.
However, model selection is still in its infancy, with grid search being
the most widely applied method. Though it works in practice, it is
computationally demanding and rather wasteful. Furthermore, it is not a
priori clear how to choose the grid.

Simplistic grid search is prone to explore large low-quality regions of
the parameter space since it does not make use of already evaluated
points to guide the search. Nested grid search improves in this aspect,
but even pure random search scales better to high-dimensional search
spaces \cite{bergstra2012random}. Direct search methods like the simplex
downhill algorithm and evolutionary optimization techniques are
applicable~\cite{glasmachers:2008c}, but they are known to require a
potentially large number of parameter evaluations.

A simple and widespread technique to speed up parameter tuning is to
subsample the data. While this technique is effective in terms of
computation time, it introduces a systematic bias since parameters are
tuned for a much less well sampled variant of the learning problem.

In this paper we focus on the key idea that relatively inaccurate models
are sufficient for identifying the well-performing parameter regime.
This introduces yet another lever that is orthogonal to both subsampling
and the search strategy in the sense that it can be combined freely with
existing techniques. This lever is the stopping criterion of the SVM
solver. 
Our approach is based on the observation that practically all SVM
training procedures in common use are based on iterative optimization
techniques with relatively low iteration cost. Hence these solvers can
easily be turned into anytime algorithms, using (processor or wall clock)
time as a stopping criterion.

Our main goal is to establish an efficient alternative to the standard
proceeding. To this end we propose to combine time-limited training with
the Efficient Global Optimization (EGO) search method~\cite{jones1998efficient}.
We explore this approach for six different types of SVM solvers on a
variety of large scale benchmark data sets.
Our results show huge and in some cases surprising differences between
solvers. They indicate that LASVM is particularly suitable for fast
model selection. We show that the time for model selection with our
method is faster than grid search by an order of a magnitude, while
loosing only little accuracy.

Independent from our research, the effect of stopping stochastic
gradient methods early was analyzed in \cite{2015arXiv150901240H}
on a theoretical level.
While our approach here can be deemed as an empirical verification
of their theory, our experiments show that this insight also holds
true for other train\-ing methods.

\section{Kernelized Support Vector Machines}
\label{section:svm}

Kernelized Support Vector Machines (SVM) \cite{cortes1995support}
are binary classifiers that use a kernel $k$ to allow for non-linear
classification decisions
$x \mapsto \sign( \langle w, \phi(x) \rangle + b)$, where $\phi$ is the
(non-linear) feature map corresponding to the kernel
$k(x, x') = \langle \phi(x), \phi(x') \rangle$ in the reproducing kernel
Hilbert space $\mathcal{H}$.
We use the RBF kernel $k(x,x') = e^{-\gamma || x-x' ||^2}$, since it
yields excellent performance and enjoys a universal approximation
property.
The SVM training problem is given by
\begin{equation}
\min_{w \in \mathcal{H}, b \in \R} \quad \frac{1}{2} ||w||^2 + C \cdot \sum_{i=1}^n \max \Big( 0, 1 - y_i \big( \langle w, \varphi(x_i) \rangle_{\mathcal{H}} + b \big) \Big),
\label{primalproblem}
\end{equation}
where $\big\{(x_1, y_1), \dots, (x_n, y_n)\big\}$ are the labeled
training points and $C > 0$ is a regularization parameter that controls
the complexity of the predictive model.

There are several methods to find a solution to~\eqref{primalproblem}.
The gold standard is Sequential Minimal Optimization (SMO), a
decomposition method solving~\eqref{primalproblem} in its dual
form \cite{platt98fast,cc01a}.
In \cite{bordes-ertekin-weston-bottou-2005} a reordering of the SMO
steps is proposed that allows for online learning. Alternatively, the
primal problem can be optimized directly, e.g., BSGD applies stochastic
gradient descent and introduces a
budget to control the (computational) complexity of the solution.
BVM and CVM solve a modified version of~\eqref{primalproblem}
with squared hinge loss. Both methods map the training problem to
the geometric problem of finding a minimal enclosing ball.

 \begin{table}[t]
 \centering
\begin{tabular}{@{}llll@{}}
\toprule
  SVM Solver  &  Method & URL  \\ 
\midrule
  LIBSVM  &  SMO  &   {\footnotesize \url{ http://www.csie.ntu.edu.tw/~cjlin/libsvm/ } }  \\ 
  BGSD  &  Stochastic Gradient  & {\footnotesize \url{ http://www.dabi.temple.edu/budgetedsvm/ } }  \\ 
  LASVM  &  Online SMO  & {\footnotesize \url{ http://leon.bottou.org/projects/lasvm } }  \\ 
  BVM/CVM &  Enclosing Ball  & {\footnotesize \url{ http://www.c2i.ntu.edu.sg/ivor/cvm.html } }  \\ 
  SVMperf  & Cutting Planes  & {\footnotesize \url{ http://svmlight.joachims.org/svm_perf.html } }  \\ 
\bottomrule
 \end{tabular}
 \caption{Overview of the applied SVM solvers. All solvers are
   implemented in C/C++, all of them can take advantage of sparsely
   represented data.}
 \label{table:Overview_of_the_solvers}
 \end{table}

%

\section{SVM Model Selection Methods}


{\bf Grid Search}: One of the basic and most often used methods to find
a good model is to solve the SVM problem on a discretized grid in the
$(C, \gamma)$ parameter space. The combination $(C, \gamma)$ with the
best cross-validation performance is eventually selected as the final
model.
Though straight-forward to use, grid search depends on the
discretization (extent and resolution of the grid), and it is not clear
how to choose it. A too small or too coarse grid might miss a good model,
while a too wide or too fine grid is wasteful.

\noindent{\bf Efficient Global Optimization (EGO)}: EGO
\cite{jones1998efficient} is a sequential, model-based (Bayesian)
optimization method. It models the error landscape with a surrogate
regression model, trained with all previous parameter evaluations,
and optimizes this model. A Kriging (Gaussian process) model is
a standard choice for the surrogate. EGO is a global optimizer. It
avoids getting stuck in local minima by optimizing the expected
improvement (EI) instead of the surrogate model's mean response.
It was applied to SVM model selection in~\cite{koch2012tuning}.

\section{Model Selection with Time-limited SVM Training}

The ultimate goal of any model selection method is to find a good model,
here, parameters $(C, \gamma)$ resulting in low generalization error.
To reach this goal we must be able to compare candidate parameters, e.g.,
with a (cross-) validation error measure. This requires training the
learning machines for each parameter setting. When stopping an SVM solver
early then we generally expect the error to be higher. However, if
stopping early results mainly in a constant shift of the error then the
error values of different parameter vectors are still in the correct
order, and the best parameters can be identified. However, it can be
expected that non-constant systematic biases exist, and that the noise
increases. The question arises whether good and bad models can be
identified early on during training or not. Our hypothesis is that this
is indeed the case.

Therefore we propose to augment each SVM solver with an additional
time-based stopping criterion. This introduces an additional parameter
into the model selection process, the stopping time $T$ controlling
the approximation quality. It needs to be set with a heuristic.

\section{Experimental Setup}

We aim to answer two questions with our experiments:

\begin{enumerate}
\item
	\textit{
	Is our model selection procedure (combining EGO with time-limited\\
	training) superior to the standard proceeding based on grid search?
	}
\item
	\textit{
	Which SVM solver is best suited for time-limited model selection?
	}
\end{enumerate}

Our method is targeted at large data sets, where model selection becomes
a computational bottleneck. Ideally, we would compare our approach to an
exhaustive grid search. However, this is computationally infeasible.
Therefore we use the results of a previous experiment~\cite{horn2014b}
based on ParEGO, a multi-criteria generalization of EGO, as a baseline
method. ParEGO optimizes contradicting goals, here accuracy and training
time, and is known to yield a very good approximation to the true Pareto
front. As such it is a much stronger baseline than grid search in terms
of training time, while yielding competitive accuracy.

All experiments were conducted on a 16-core CPU with 64 GB of RAM.%
\footnote{This is a different hardware setup than the high performance
  cluster used in the ParEGO experiments. This is compliance with our
  goal to show that an extensive model search can be mimicked on a
  commodity workstation. Furthermore, the difference in speed per core
  of both setups can be expected to be a small factor and hence does not
  affect our main result.}
We consider the following data sets:
arthrosis, aXa, cod-rna, covtype, ijcnn1, mnist, poker, protein, shuttle, spektren, vehicle and wXa.
All data sets have been split randomly into training, validation, and
test sets with a ratio of 2:1:1.

A user-adjustable time constraint%
\footnote{We make all our modified software packages, the experiment
  with all details and all results publicly available at
  {\url{https://www.github.com/aydindemircioglu/TLEGO}}.}
was added to the original software packages,
see Table~\ref{table:Overview_of_the_solvers}.
We left all tunable parameters of the solvers at their defaults,%
\footnote{Refer to our repository and also to \cite{horn2014b} for more details.}
except for the budget of BSGD, which we fixed to 2048. For EGO we used
the implementation in mlrMBO.%
\footnote{\url{https://github.com/berndbischl/mlrMBO}}
To speed up EGO, we ran a parallel version, in which
we used the lower confidence bound criterion (LCB) instead of the usual expected
improvement. In order to propose multiple points from the model for parallel
evaluation, we sampled multiple values of the $\lambda$ parameter of the LCB criterion
and optimized each $\lambda$-LCB-function independently. Further details concerning EGO parallelization can be found in \cite{bischl2014moimbo}.
Similar to
\cite{horn2014b}, $C$ and $\gamma$ were constrained to the range
$[2^{-15}, 2^{15}]$, which is frequently used for a thorough grid search.
The initial design for each run was chosen to consist of $20$ points,
and $10$ sequential iterations with $20$ points proposed in parallel
were performed.

We fixed the training time-limit to $T = 2^{\log_{10}(n) + 1}$ seconds,
which we have chosen heuristically.
For each point proposed by EGO a solver was run for $T$ seconds on the
training set and afterwards the resulting model was evaluated 
on the corresponding
validation set. The final model was trained with the best parameters
found with a time-limit of 8 hours and evaluated on the test set.

Statistical significance was tested with a Friedman test~\cite{demvsar2006statistical},
where the null hypothesis is that there is no difference between the
classifiers.

\section{Results and Discussion}


\begin{figure}[!bt]
	\hspace{-0.5cm} 
	\small
	\setlength{\tabcolsep}{0cm}
	\begin{minipage}{0.5\linewidth}
	\includegraphics[width=1.0 \textwidth]{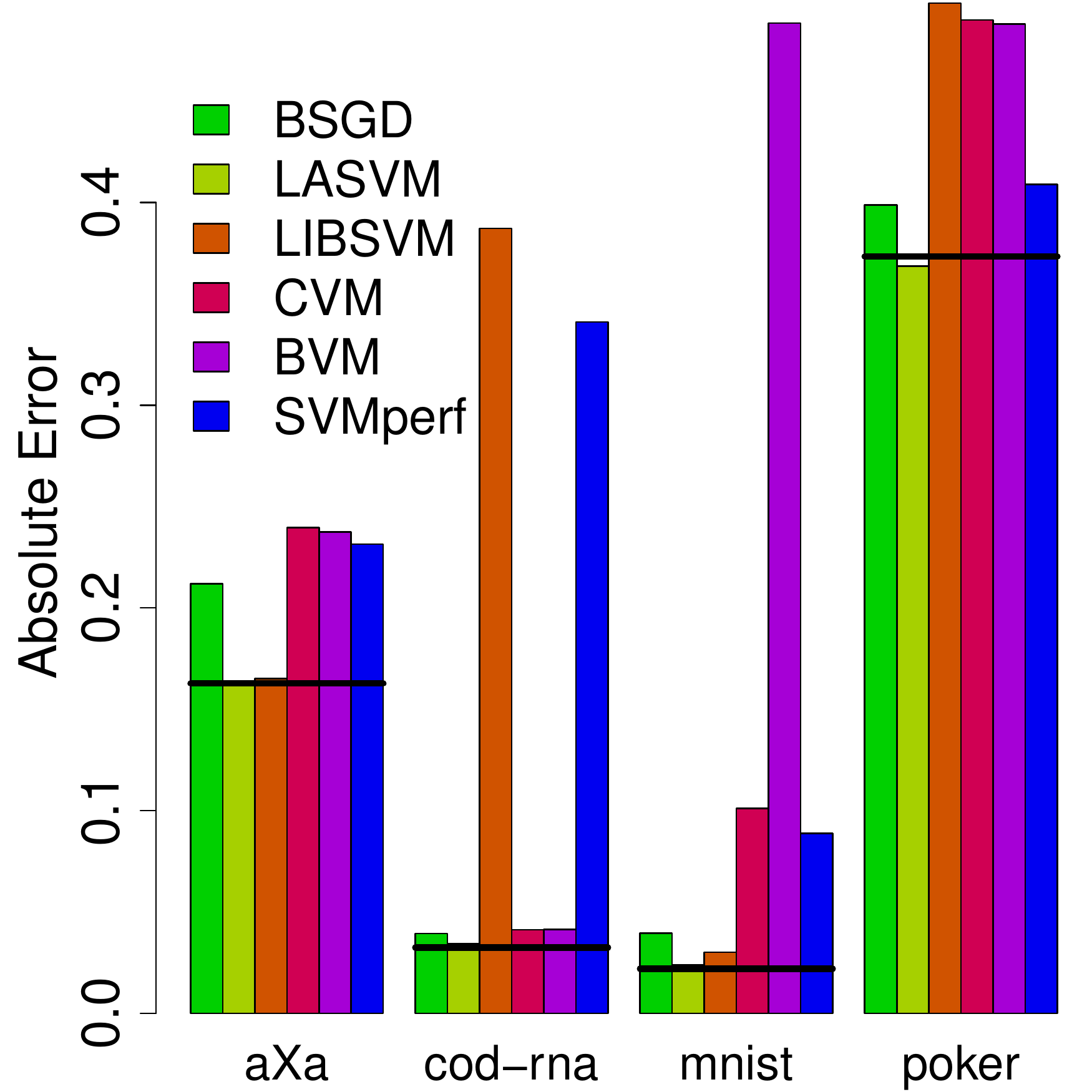}
\caption{Absolute Error rates (see text)} 
\label{plot:error}
	\end{minipage}%
	\hspace{0.4cm}
	\setlength{\tabcolsep}{0cm}
	\begin{minipage}{.5\linewidth}
	\includegraphics[width=1.0 \textwidth]{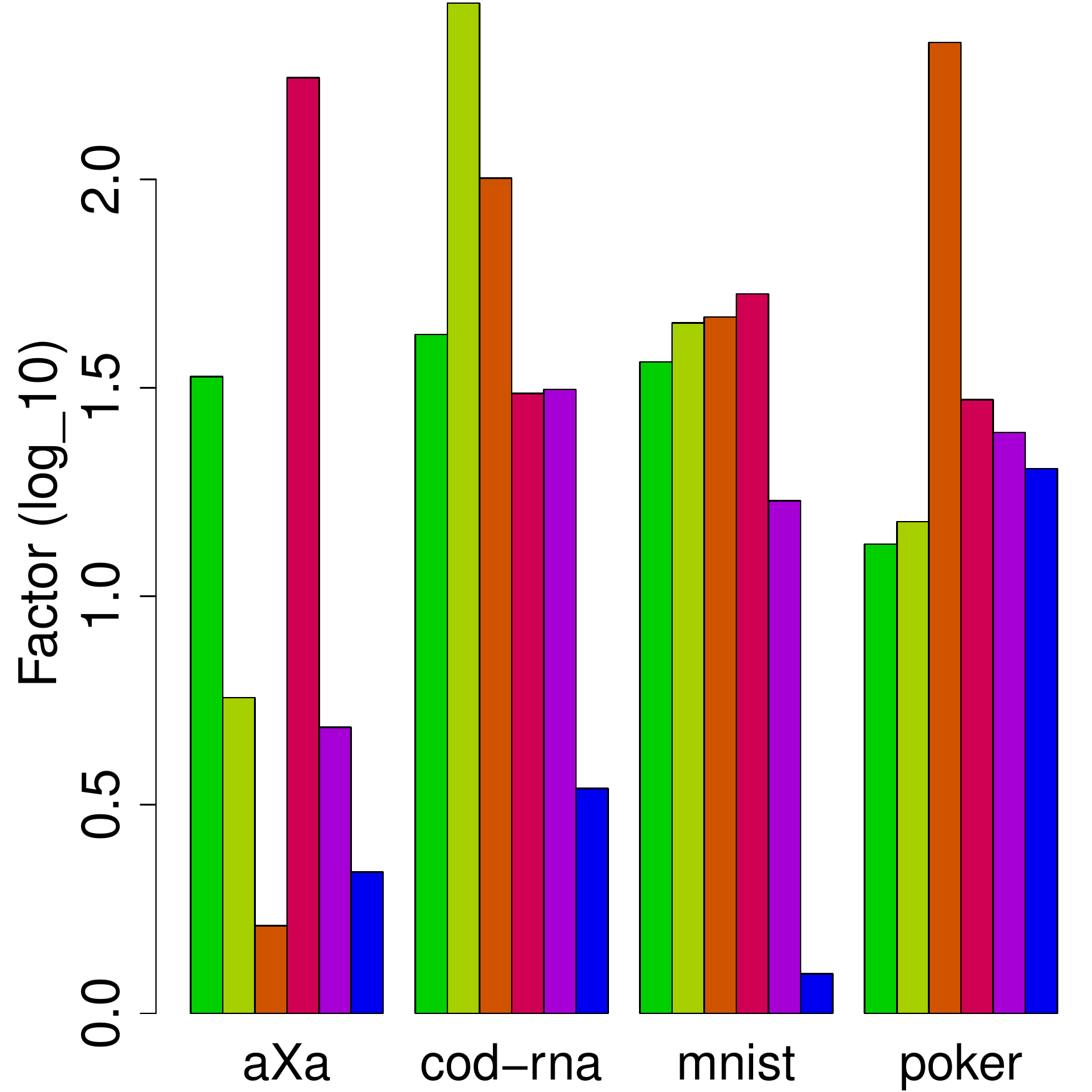}
\caption{Relative Timings (see text)} 
\label{plot:timing}
	\end{minipage} 
\end{figure}

Figure~\ref{plot:error} shows the error rates (lower is better) achieved
by the final model when performing time limited model selection, for all
six solvers. The black bars indicate the performance of the baseline
experiment.%
\footnote{
Due to space constraints the figure shows only representative results
for four data sets. All results are available at
{\url{http://largescalesvm.de/tlego/}}.
}
Figure~\ref{plot:timing} displays the corresponding runtime: the sum of
training and validation times during model selection and the training
time of the final model. The values are the order of magnitude (base-10
logarithm) of the factor by which our method is faster than the baseline
(higher is better).

%

We start with our second question. From Figure \ref{plot:error} it is
apparent that apart from LASVM the comparability of models suffers from
time-limited training on at least one data set. None of the other
solvers performs well consistently.
Indeed, the Friedman test indicates a highly significant difference
between the solvers ($p < 10^{-7}$). A post-hoc test using Holm's 
method~\cite{demvsar2006statistical}
comparing ParEGO to all other solvers, shows that
there is a significant difference between ParEGO and LASVM in one group,
and all other solvers.
Therefore, our method only applies to LASVM. Naively we would have
expected LIBSVM to perform very similar to LASVM as both solve the dual
problem with SMO, but this seems not to be true, as the results are not
decisive enough. We suspect that LIBSVM's models are only comparable in
a late convergence phase. LASVM's focus on online optimization helps to
produce better (comparable) models from the very beginning.

Focusing only on LASVM, Figure~\ref{plot:timing} shows that our method
yields a major speed up over ParEGO, and therefore over grid search, of
roughly one to two orders of magnitudes. This gives a strong affirmative
answer to our first question.

\section{Conclusions}

We show that a simple modification of the
stopping criterion by limiting the time spent on training 
together with the EGO search strategy
can be used to speed up model selection without loosing  accuracy
significantly by more than an order of magnitude with respect to ParEGO
and thus to grid search.
This result depends heavily on the SVM solver, and holds true only for LASVM.
On a theoretical level much needs to be done.
There is no explanation yet for the different behavior of the SVM solvers.
Also it is unclear how to choose the time limit. We have not explored
a more dynamical method, e.g., making  the time spent on training
dependent on the relative performance and using a more refined search
once the relevant region is determined.
Finally, it is interesting to analyze how our stopping criterion
interacts with subsampling, which is an alternative to speeding up model
selection, and  to apply this principle to other, more sophisticated
model search methods.



\begin{footnotesize}


\bibliographystyle{unsrt}
\bibliography{references}

\end{footnotesize}


\end{document}